\documentclass[runningheads]{llncs}
\usepackage[T1]{fontenc}
\usepackage{graphicx}
\usepackage{amsmath,amssymb,amsfonts}
\usepackage{booktabs}
\usepackage{bm}
\usepackage{multirow}
\usepackage{subfig}
\usepackage{threeparttable}
\usepackage[table]{xcolor}

\begin{document}

\title{Empirical Boundaries of Reservoir Computing for Few-Shot Sequence Pattern Tasks}

\titlerunning{Empirical Boundaries of RC for Few-Shot Sequence Patt. Tasks}

\author{Mario Mallea\orcidID{0009-0005-8295-088X} \and
Àngela Nebot\orcidID{0000-0002-4621-8262} \and
Francisco Mugica\orcidID{0000-0003-2843-0427}}
\authorrunning{M. Mallea et al.}

\institute{Universitat Politècnica de Catalunya - BarcelonaTech, Soft Computing Research Group at the Intelligent Data Science and Artificial Intelligence Research Center, Barcelona, Spain. \email{mario.carlos.mallea@upc.edu}}

\maketitle  
\vspace{-0.1 cm}
\begin{abstract}
Efficient sequence modeling remains a fundamental challenge in machine learning. Despite the success of Transformers, recurrent neural networks (RNNs) have recently re-emerged as efficient alternatives. Particularly, Echo State Networks (ESNs) offer a compelling framework of randomized RNNs that have been extensively studied for time-series forecasting and dynamical systems. This paper presents an empirical analysis of distributional, architectural, and optimization variants of ESNs evaluated on few-shot sequence-to-sequence (seq2seq) pattern tasks. These synthetic tasks utilize discrete symbol sequences, characterized by delays, inactive phases, and triggered markers to isolate and evaluate memory and pattern recognition capabilities. The minimality of these tasks is crucial to understand the strengths and limitations of machine learning models. Furthermore, when the input signal utilizes a one-hot encoding, ESN dynamics are governed strictly by their weight distributions, offering an ideal lens through which to study reservoir computing. 

Our findings map the performance boundaries of ESNs relative to state-of-the-art (SOTA) models. We show that preserving a linear state regime is vital for memory tasks, such as pattern completion or simple copy. We propose a data-dependent modulation mechanism for nonlinear input injection into the reservoir (G-ESN). G-ESN functions as a statistical filter capable of solving tasks that conventional ESNs cannot (the adding problem and bracket matching). Finally, for Deep ESNs, we evaluate a shared readout mechanism across layers that outperforms conventional concatenated readouts on structural tasks, including selective copy, sorting problem, and bracket matching.

\keywords{Reservoir Computing \and Echo State Network \and Sequential Modeling.}
\end{abstract}

\section{Introduction}

Foundation models based on the Transformer architecture have advanced artificial intelligence. However, the quadratic computational complexity of self-attention restricts efficient inference over long sequences and limits deployment in resource-constrained environments. Moreover, the attention mechanism operates primarily as an associative short-term memory, often lacking the persistent state dynamics needed to track long-term dependencies. In contrast, RNNs exhibit linear scaling with respect to sequence length, providing faster inference times, even for modest sequence lengths \cite{orvieto2023resurrecting,liu2020mkdmultitaskknowledgedistillation}. Recently, architectures like the Selective State Space model (S6/Mamba) \cite{gu2024mamba}, Titans \cite{behrouz2024titans}, and modern linear recurrent models \cite{dao2024transformers,orvieto2023resurrecting} have successfully approximated Transformers’ language modeling capabilities while addressing the efficiency and memory concerns.

Efficiency and memory capacity are well understood properties of ESNs \cite{jaeger2001short}. An ESN delegates temporal processing to a random dynamical system known as the reservoir, typically operating near the edge of chaos. While ESNs have proven successful in domains such as ensemble forecasting \cite{MALLEA2026110854}, time-series classification \cite{li2025novel}, wind power forecasting \cite{lopez2018wind,yuzgec2024novel} and battery health prognosis \cite{huang2025state,MALLEA2025100577}, they remain understudied in the context of discrete sequence modeling. Appendix \footnote{https://github.com/MariodotR/ESN-on-cognitive-tasks.git} 
Sec. 1 provides an overview of how randomness and reservoir computing have been utilized to study and assess sequence modeling.

Theoretically, purely linear ESNs possess a maximum memory capacity equal to the reservoir size \cite{ballarin2024memory}. To extend this capacity, various ESNs have been proposed, including the LeakyESN \cite{leaky}, Deep ESNs \cite{gallicchio2017deep} with concatenated layer readouts, and gated reservoir variants inspired by GRU \cite{di2020gated} or LSTM \cite{lun2024long} models. More recently, the Echo State Transformer (EST) \cite{bendi2025echo} attempted to bridge the gap between reservoirs and attention mechanisms by integrating parallel reservoirs as a working memory core within self-attention layers. 

Despite these developments, critical questions persist: Which sequence tasks are solvable using ESNs, and which are open challenges? How do their boundaries differ from SOTA architectures? To the best of our knowledge, we present the first work to evaluate few-shot seq2seq pattern tasks on architectural, distributional, and optimization variants of ESNs and benchmarking them directly against SOTA models. Specifically, our primary contributions are:

\begin{itemize}
\item We mathematically formalize and categorize the tasks within the STREAM benchmark \cite{bendi2025echo}, adding to the benchmark the results for ESN variants and the S6 architecture. We map the trade-offs of reservoir computing against fully trained models across specific task archetypes: state tracking and relational routing. Notably, the latter tasks (e.g selective copy and sorting problem) remain challenging for ESNs.
\item We show that input weight initialization is critical when processing one-hot encoded data. Conventional independent and identically distributed (IID) initialization causes rapid state-information collapse. Conversely, we argue that orthonormal (ON) initialization enables nonlinear activations to function within a stable, quasi-linear regime. This allows for near-perfect performance on state tracking tasks, such as simple copy and discrete pattern completion, which SOTA models fail to achieve.
\item We propose a generalization of the LeakyESN into a data-dependent, adaptive leaky integrator (G-ESN) to extend reservoir capacity. We show that standard reservoir states are insufficient for tasks like the adding problem and bracket matching, and we argue how G-ESN is able to implement a filter to repair that. Indeed, the proposed G-ESN is the only model to achieve near-zero error on bracket matching.
\item We evaluate a shared readout mechanism across stacked layers in Deep ESNs, contrasting it with traditional layer concatenation. We demonstrate that tasks such as sequential MNIST, the adding problem, selective copy, and bracket matching, can benefit from the structural regularization imposed by a shared readout rather than concatenated feature selection.
\end{itemize}

\section{Echo State Networks}

Let $u_t \in \mathbb{R}^{d_i}$ represent the input vector and $h_t \in \mathbb{R}^{d_r}$ denote the reservoir state at time step $t$. The state transition dynamics are formalized as:
\begin{equation}\label{eq:general_update}
    h_{t+1} = (\mathbf{1} - \alpha_t) \odot h_{t} + \alpha_t \odot \phi(p_t),
\end{equation}
where $\mathbf{1} \in \mathbb{R}^{d_r}$, $\alpha_t \in [0, 1]^{d_r}$ represents the leaky integration vector, and $\phi$ is the activation function applied to the pre-activation vector $p_t= u_t W_{in} + h_t W_{res}$. $W_{in} \in \mathbb{R}^{d_i \times d_r}$ represents the input weight matrix and $W_{res} \in \mathbb{R}^{d_r \times d_r}$ denotes the recurrent reservoir weight matrix. To guarantee the echo state property, the spectral radius is strictly bounded via $\rho(W_{res}) < 1$. We construct $W_{res}$ as a scaled orthogonal matrix, $W_{res} = 0.9 Q / \rho(Q)$, utilizing the QR decomposition $QR = \text{qr}(M)$, where $M_{ij} \sim \mathcal{U}(-1, 1)$ \cite{ON_reservoir}. This unitary structure preserves the norm of the state vector over time while maximizing the network's effective memory capacity.

Regarding the input matrix $W_{in}$, we analyze two distinct initialization regimes: \textbf{IID:} a high-entropy projection where entries are drawn independently from a uniform distribution, $W_{in_{i,j}} \sim \mathcal{U}(-1, 1)$. \textbf{ON:} an isotropic projection with orthogonal directions and identical gains, where $W_{in_{i,j}} \sim \mathcal{N}(0, 1)$ and its columns are transformed into an orthonormal basis via QR decomposition, scaled by $1/\sqrt{d_i}$.
\vspace{-0.5 cm}
\subsubsection{Architectures.} Different reservoir models are derived from \eqref{eq:general_update}:

\begin{itemize}
    \item \textbf{LESN:} Defines a purely linear dynamical system by setting $\phi(z) = z$ and $\alpha_t = \mathbf{1}$ for all $t$. This variant serves as a baseline to isolate the exact impact of nonlinear feature expansion.
    \item \textbf{ESN:} Represents the canonical reservoir model with a nonlinear state transformation at each step, using $\phi(z) = \tanh(z)$ and $\alpha_t = \mathbf{1}$ for all $t$.
    \item \textbf{LeakyESN:} Formulates a convex combination of the nonlinear state update and a historical skip connection by setting $\phi(z) = \tanh(z)$ and $\alpha_t = \alpha \odot \mathbf{1}$, where $\alpha \in (0, 1]$ is a fixed hyperparameter.
    \item \textbf{G-ESN:} Introduces an adaptive, data-dependent leaky integrator where $\phi(z) = \tanh(z)$ and the leak rate is modulated via: $\alpha_t = \sigma(W_{gate} u_t + b_{gate})$, where $\sigma(\cdot)$ is the logistic sigmoid function, $W_{gate} \in \mathbb{R}^{d_r \times d_i}$ is a random matrix, and $b_{gate} = b \odot \mathbf{1}$ represents a uniform bias hyperparameter with $b\in\mathbb{R}$. 
    
    To illustrate this gating behavior, let $g_t = W_{gate} u_t$ denote the unbiased gate pre-activation. On the one hand, with a baseline bias of $b = -3$, the pre-activation yields $g_{\text{symbol}} + b \in [-3, -2]$ during typical sequence processing steps, but increases to $g_{\text{symbol}} + g_{\text{mask}} + b \in [-3, -1]$ when a marked token appears. Consequently, any reservoir unit satisfying $\alpha_t > \sigma(-2)$ explicitly flags the masked token, incorporating a higher proportion of the nonlinear update. On the other hand, for two contrasting symbols ($A$ and $B$) under a contrastive initialization where $W_{gate}[:,A] > 0 > W_{gate}[:,B]$ and $b=0$, the G-ESN modulates state preservation such that $\sigma(g_B) < 0.5 < \sigma(g_A)$.
    \item \textbf{Deep ESN:} Consists of $L$ hierarchically stacked reservoir layers, where the state of layer $l$, denoted as $h^{(l)}_t$, is driven by the state of the preceding layer: $h^{(l)}_t = \tanh(W_{in}^{(l)} i^{(l)}_t + W_{res}^{(l)} h^{(l)}_{t-1})$, where $i^{(1)}_t = u_t$ and $i^{(l)}_t = h^{(l-1)}_t$ for $l > 1$. Each layer's weight matrices are generated as independent random realizations of the selected initialization strategy.
\end{itemize}
\vspace{-0.5 cm}
\subsubsection{Optimization.} We apply linear regression to map the hidden features to the target labels. The readout weights $W_{out} \in \mathbb{R}^{d_r \times d_o}$ are trained via Ridge regression (Tikhonov regularization) to minimize the mean squared error across the designated evaluation time steps $T$:
\begin{equation}
    W_{out} = \underset{W}{\operatorname{argmin}} \| Y - H W \|_F^2 + \lambda \| W \|_F^2,
\end{equation}
where $H \in \mathbb{R}^{N \cdot T \times d_r}$ represents the flattened matrix of reservoir states across $N$ training samples at $T$ evaluation times, $Y \in \mathbb{R}^{N \cdot T \times d_o}$ denotes the corresponding target matrix (seq2seq tasks), and $\lambda$ is the regularization hyperparameter. This optimization problem yields a closed-form solution:
\begin{equation}\label{eq:pseudoinverse}
    W_{out} = (H^T H + \lambda I)^{-1} H^T Y.
\end{equation}

For Deep ESNs containing $L$ layers, we implement and compare two distinct readout mechanisms:
\begin{itemize}
    \item \textbf{Concatenated:} The conventional approach performs linear regression from the concatenation of all hidden layer states, $H = [h^{(1)}_t \mid \dots \mid h^{(L)}_t] \in \mathbb{R}^{N \cdot T \times d_r \cdot L}$ and $W_{out} \in \mathbb{R}^{d_r \cdot L \times d_o}$, allowing the readout to perform feature selection simultaneously across multiple abstraction timescales. 

    $$\hat{d_r}= d_r \cdot L 
    $$
    \item \textbf{Shared:} We propose constraining the readout weight matrix $W_{out} \in \mathbb{R}^{d_r \times d_o}$ to be shared invariants across all hidden layers, solving the joint minimization problem:
    \begin{equation}
        W_{out} = \underset{W}{\operatorname{argmin}} \sum_{i=1}^{L} \| Y - H^{(i)} W \|_F^2 + \lambda \| W \|_F^2.
    \end{equation}
    Enforcing shared weights across layers acts as a structural regularizer, compelling the optimization procedure to discover task-invariant manifolds across depths rather than relying on wide feature selection. This formulation also has a closed-form solution \cite{MALLEA2026110854}.
\end{itemize}

\section{Experimental Setting}

\subsubsection{Tasks.} We evaluate all ESN variants on the STREAM benchmark, strictly adhering to the evaluation protocols (including identical seeds and data partitions) established by \cite{bendi2025echo}. STREAM comprises twelve distinct sequence tasks designed to evaluate a model's capacity to extract seq2seq temporal patterns from a limited number of examples, testing few-shot generalization. We categorize these twelve tasks into two fundamental groups based on the underlying mathematical mapping between the input and target sequences (detailed formal descriptions are available in Appendix Sec. 3). The first group are state tracking tasks, where the target sequence is a linear combination or a counter-based function of the input history. For the discrete domain, these include simple copy (Copy), adding problem (Adding), bracket matching (Bracket), discrete pattern completion (Disc. patt.), and discrete postcasting (Disc. post.).
The second group consists of relational routing tasks, which are data-dependent and content-aware sequence problems requiring structural routing. These include selective copy (S-copy), sorting problem (Sorting), and Sequential MNIST (SMNIST). 

Discrete and classification tasks are evaluated using error rate ($1 - \text{accuracy}$). The remaining four tasks are continuous problems: sinus forecasting (Sinus), chaotic forecasting (Chaotic), continuous pattern completion (Cont. patt.), and continuous postcasting (Cont. post.), which are evaluated using mean squared error.

Models are evaluated following a trigger token and a delay period, testing long-term information retention. For example, the copy task uses a sequence length of 50 steps across 4 dimensions: the first 22 steps contain a 3-symbol sequence, followed by 5 delay steps (zeros), a trigger marker (the fourth dimension), and 22 subsequent zeros. The target requires reproducing the original 22-step sequence during the final zero-input phase. Then, note that the seq2seq formulation forces the model to rely purely on its autonomous hidden dynamics or the fixed points of the $\tanh$ activation. Fig. 1 in Appendix illustrates this seq2seq formulation for the selective copy task.
\vspace{-0.5 cm}
\subsubsection{Implementation details.} To ensure architectural robustness, all experiments were conducted across 5 random seeds following \cite{bendi2025echo}. We report the mean, standard deviation, and minimum error. Tab.~\ref{tab:hyperparams} outlines the hyperparameter grid-search configurations. Our code is publicly available.
\vspace{-0.5 cm}
\begin{table}[h!]
\centering
\caption{Validation experiments for ESN variants. Best $\lambda$ found for ESN was transferred to LeakyESN and G-ESN.}
\label{tab:hyperparams}
\begin{tabular}{l c c c }
\hline
Model & Hyperparameter  & Grid Search & Results (Appendix) \\
\hline
ESN and LESN & $\lambda$ &  $[0.0001,0.001,0.01,0.1,1,10,100]$ &  Tab. 2 \\
Deep ESN variants & $\lambda$ & $[0.0001,0.001,0.01,0.1,1,10,100]$  & Tab. 3\\
LeakyESN &  $\alpha$ & $ [0.01,0.1,0.3,,.05,0.7,0.9]$ & Tab. 4 \\
G-ESN & $b$ & $ [0,-1,-3,-5,-7,-10]$ & Tab. 5\\
\hline
\end{tabular}
\end{table}

We benchmark our reservoir methods against the best models of \cite{bendi2025echo}: LSTM, GRU, Transformer, and EST, with parameter scales ranging from 1K to 1M trainable weights. In addition, we trained the S6 \cite{gu2024mamba}, a candidate to replace Transformers as backbone. S6 is a linear-time RNN trained with backpropagation through time, that introduces data-dependent gating to selectively reset latent states based on current inputs. This ability was demonstrated on S-copy, improving Transformers with the advantage of the efficient autoregressive inference. Training details are detailed in Appendix Sec. 2.

The reservoir size was fixed to $d_r = 10,000 / (L \cdot d_o)$, setting $L=1$ for single-layer ESN variants, and $L=3$ for the Deep ESN with concatenated readout (DESN C.) for a fair comparison with ESNs. Thus, each layer has one-third of the dimensions, but the number of trainable parameters remains the same. We also evaluate two shared readout configurations: DESN S. and DESN S.F. The former uses the same layers of DESN C., having a one-third of the trainable parameters. The latter takes advantage of the readout and uses each layer with full reservoir size (as single-layer ESN variants), matching the trainable parameters of DESN C. and single-layer ESN variants.

\section{Results and Discussion}

Tables \ref{tab:ablation_complete1}, \ref{tab:ablation_complete2}, and \ref{tab:ablation_complete3} report the performance of our ESN variants for the twelve tasks, the mean, standard deviation, and minimum errors are shown among five seeds as \cite{bendi2025echo}. Best mean and minimum errors are in bold per task, and $*$ denotes a statistically significant difference ($p-\text{value} < 0.05$ of an unequal variances t-test) in favor of the best initialization per model and task.
\vspace{-0.7 cm}
\begin{table*}[h]
\centering
\caption{Reservoir computing performance for the tasks: Sinus, Chaotic, Disc. post., and Cont. post. }
\label{tab:ablation_complete1}
\resizebox{\textwidth}{!}{%
\begin{tabular}{ll c c c c }
\toprule
\textbf{Model} & \textbf{Init} & \textbf{Sinus} & \textbf{Chaotic} & \textbf{Disc. post.} & \textbf{Cont. post.} \\
\midrule
\multirow{2}{*}{LESN} & IID & 0.001 $\pm$ 0.000 (0.001) & 0.042 $\pm$ 0.026 (0.021) & 0.000 $\pm$ 0.000 \textbf{(0.000)} & \textbf{0.000} $\pm$ 0.000 \textbf{(0.000)} \\
 & ON & 0.001 $\pm$ 0.001 (0.001) & 0.044 $\pm$ 0.025 (0.020) & \textbf{0.000} $\pm$ 0.000 \textbf{(0.000)} & 0.000 $\pm$ 0.000 (0.000) \\
\midrule
\multirow{2}{*}{ESN} & IID & \textbf{0.000} $\pm$ 0.000 (0.000) & \textbf{0.011} $\pm$ 0.000 \textbf{(0.011)} & 0.000 $\pm$ 0.000 \textbf{(0.000)} & 0.000 $\pm$ 0.000 (0.000) \\
 & ON & 0.001 $\pm$ 0.001 (0.001) & 0.044 $\pm$ 0.025 (0.020) & \textbf{0.000} $\pm$ 0.000 \textbf{(0.000)} & 0.000$^{*}$ $\pm$ 0.000 (0.000) \\
\midrule
\multirow{2}{*}{G-ESN} & IID & 0.001 $\pm$ 0.000 (0.001) & 0.368 $\pm$ 0.175 (0.024) & 0.443 $\pm$ 0.175 (0.092) & 0.096 $\pm$ 0.002 (0.094) \\
 & ON & 0.003 $\pm$ 0.003 (0.001) & 0.117$^{*}$ $\pm$ 0.101 (0.036) & 0.433 $\pm$ 0.049 (0.337) & 0.101 $\pm$ 0.004 (0.098) \\
\midrule
\multirow{2}{*}{LeakyESN} & IID & 0.000 $\pm$ 0.000 \textbf{(0.000)} & 0.025 $\pm$ 0.004 (0.017) & 0.085 $\pm$ 0.068 \textbf{(0.000)} & 0.000 $\pm$ 0.000 (0.000) \\
 & ON & 0.001 $\pm$ 0.001 (0.001) & 0.069 $\pm$ 0.042 (0.022) & \textbf{0.000} $\pm$ 0.000 \textbf{(0.000)} & 0.000$^{*}$ $\pm$ 0.000 (0.000) \\
\midrule
\multirow{2}{*}{DESN S.} & IID & 0.001$^{*}$ $\pm$ 0.000 (0.001) & 0.029$^{*}$ $\pm$ 0.003 (0.024) & 0.000 $\pm$ 0.000 \textbf{(0.000)} & 0.009 $\pm$ 0.000 (0.009) \\
 & ON & 0.001 $\pm$ 0.000 (0.001) & 0.036 $\pm$ 0.005 (0.030) & \textbf{0.000} $\pm$ 0.000 \textbf{(0.000)} & 0.000$^{*}$ $\pm$ 0.000 (0.000) \\
\midrule
\multirow{2}{*}{DESN S.F.} & IID & 0.001 $\pm$ 0.000 (0.001) & 0.027$^{*}$ $\pm$ 0.002 (0.023) & \textbf{0.000} $\pm$ 0.000 \textbf{(0.000)} & 0.009 $\pm$ 0.000 (0.009) \\
 & ON & 0.002 $\pm$ 0.002 (0.001) & 0.040 $\pm$ 0.008 (0.029) & \textbf{0.000} $\pm$ 0.000 \textbf{(0.000)} & 0.000$^{*}$ $\pm$ 0.000 (0.000) \\
\midrule
\multirow{2}{*}{DESN C.} & IID & 0.001$^{*}$ $\pm$ 0.000 (0.001) & 0.033$^{*}$ $\pm$ 0.003 (0.029) & \textbf{0.000} $\pm$ 0.000 \textbf{(0.000)} & 0.006 $\pm$ 0.000 (0.006) \\
 & ON & 0.002 $\pm$ 0.000 (0.001) & 0.043 $\pm$ 0.006 (0.036) & \textbf{0.000} $\pm$ 0.000 \textbf{(0.000)} & 0.000$^{*}$ $\pm$ 0.000 (0.000) \\
\bottomrule
\end{tabular}%
}
\end{table*}

\begin{table*}[h]
\centering
\caption{Reservoir computing performance for the tasks: Disc. patt., Cont. patt., Bracket, and Copy.}
\label{tab:ablation_complete2}
\resizebox{\textwidth}{!}{%
\begin{tabular}{ll c c c c }
\toprule
\textbf{Model} & \textbf{Init} & \textbf{Disc. patt.} & \textbf{Cont. patt.} & \textbf{Bracket} & \textbf{Copy} \\
\midrule
\multirow{2}{*}{LESN} & IID & \textbf{0.056} $\pm$ 0.006 (0.050) & 0.050 $\pm$ 0.003 (0.044) & 0.400 $\pm$ 0.102 (0.200) & \textbf{0.000} $\pm$ 0.000 \textbf{(0.000)} \\
 & ON & 0.056 $\pm$ 0.009 \textbf{(0.046)} & 0.050 $\pm$ 0.004 (0.044) & 0.388 $\pm$ 0.085 (0.220) & \textbf{0.000} $\pm$ 0.000 \textbf{(0.000)} \\
\midrule
\multirow{2}{*}{ESN} & IID & 0.076 $\pm$ 0.008 (0.069) & \textbf{0.023}$^{*}$ $\pm$ 0.002 \textbf{(0.021)} & 0.392 $\pm$ 0.028 (0.340) & 0.400 $\pm$ 0.005 (0.394) \\
 & ON & 0.057$^{*}$ $\pm$ 0.009 \textbf{(0.046)} & 0.048 $\pm$ 0.004 (0.044) & 0.396 $\pm$ 0.066 (0.270) & \textbf{0.000}$^{*}$ $\pm$ 0.000 \textbf{(0.000)} \\
\midrule
\multirow{2}{*}{G-ESN} & IID & 0.131 $\pm$ 0.016 (0.111) & 0.042$^{*}$ $\pm$ 0.004 (0.037) & 0.256 $\pm$ 0.254 \textbf{(0.040)} & 0.471 $\pm$ 0.007 (0.465) \\
 & ON & 0.124 $\pm$ 0.050 (0.071) & 0.049 $\pm$ 0.003 (0.043) & \textbf{0.052} $\pm$ 0.015 \textbf{(0.040)} & 0.430$^{*}$ $\pm$ 0.012 (0.418) \\
\midrule
\multirow{2}{*}{LeakyESN} & IID & 0.095 $\pm$ 0.008 (0.081) & 0.025$^{*}$ $\pm$ 0.002 (0.023) & 0.302 $\pm$ 0.104 (0.120) & 0.464 $\pm$ 0.007 (0.453) \\
 & ON & 0.056$^{*}$ $\pm$ 0.008 (0.049) & 0.053 $\pm$ 0.004 (0.045) & 0.374 $\pm$ 0.079 (0.280) & 0.055$^{*}$ $\pm$ 0.007 (0.046) \\
\midrule
\multirow{2}{*}{DESN S.} & IID & 0.145 $\pm$ 0.007 (0.139) & 0.039 $\pm$ 0.003 (0.036) & 0.402 $\pm$ 0.050 (0.310) & 0.499 $\pm$ 0.012 (0.488) \\
 & ON & 0.058$^{*}$ $\pm$ 0.009 (0.049) & 0.043 $\pm$ 0.003 (0.038) & 0.408 $\pm$ 0.098 (0.220) & 0.001$^{*}$ $\pm$ 0.001 \textbf{(0.000)} \\
\midrule
\multirow{2}{*}{DESN S.F.} & IID & 0.141 $\pm$ 0.007 (0.134) & 0.035 $\pm$ 0.002 (0.033) & 0.378 $\pm$ 0.069 (0.300) & 0.479 $\pm$ 0.007 (0.473) \\
 & ON & 0.060$^{*}$ $\pm$ 0.007 (0.047) & 0.042 $\pm$ 0.005 (0.037) & 0.388 $\pm$ 0.072 (0.250) & \textbf{0.000}$^{*}$ $\pm$ 0.000 \textbf{(0.000)} \\
\midrule
\multirow{2}{*}{DESN C.} & IID & 0.153 $\pm$ 0.008 (0.143) & 0.037 $\pm$ 0.002 (0.033) & 0.386 $\pm$ 0.079 (0.280) & 0.473 $\pm$ 0.006 (0.463) \\
 & ON & 0.057$^{*}$ $\pm$ 0.008 (0.047) & 0.041 $\pm$ 0.002 (0.036) & 0.404 $\pm$ 0.049 (0.310) & 0.000$^{*}$ $\pm$ 0.000 \textbf{(0.000)} \\
\bottomrule
\end{tabular}%
}
\end{table*}

\begin{table*}[h]
\centering
\caption{Reservoir computing performance for the tasks: S-copy, Adding, Sorting, and SMNIST.}
\label{tab:ablation_complete3}
\resizebox{\textwidth}{!}{%
\begin{tabular}{ll c c c c }
\toprule
\textbf{Model} & \textbf{Init} & \textbf{S-Copy} & \textbf{Adding} & \textbf{Sorting} & \textbf{SMNIST} \\
\midrule
\multirow{2}{*}{LESN} & IID & 0.645 $\pm$ 0.021 (0.622) & 0.722 $\pm$ 0.025 (0.690) & 0.578 $\pm$ 0.012 (0.560) & 0.436 $\pm$ 0.081 (0.290) \\
 & ON & 0.645 $\pm$ 0.029 (0.616) & 0.720 $\pm$ 0.018 (0.690) & 0.568 $\pm$ 0.008 (0.557) & 0.452 $\pm$ 0.067 (0.320) \\
\midrule
\multirow{2}{*}{ESN} & IID & 0.634 $\pm$ 0.026 (0.594) & 0.762 $\pm$ 0.075 (0.670) & 0.601 $\pm$ 0.019 (0.579) & 0.622 $\pm$ 0.028 (0.590) \\
 & ON & 0.642 $\pm$ 0.031 (0.612) & 0.726 $\pm$ 0.015 (0.710) & 0.568$^{*}$ $\pm$ 0.009 (0.555) & 0.466$^{*}$ $\pm$ 0.050 (0.370) \\
\midrule
\multirow{2}{*}{G-ESN} & IID & \textbf{0.429} $\pm$ 0.014 \textbf{(0.410)} & \textbf{0.006}$^{*}$ $\pm$ 0.005 \textbf{(0.000)} & 0.518 $\pm$ 0.034 (0.487) & 0.448 $\pm$ 0.020 (0.420) \\
 & ON & 0.436 $\pm$ 0.015 (0.414) & 0.452 $\pm$ 0.186 (0.250) & 0.504 $\pm$ 0.011 (0.491) & \textbf{0.348}$^{*}$ $\pm$ 0.043 (0.280) \\
\midrule
\multirow{2}{*}{LeakyESN} & IID & 0.449$^{*}$ $\pm$ 0.005 (0.440) & 0.316$^{*}$ $\pm$ 0.074 (0.240) & \textbf{0.477} $\pm$ 0.027 \textbf{(0.433)} & 0.414 $\pm$ 0.079 (0.320) \\
 & ON & 0.618 $\pm$ 0.013 (0.596) & 0.698 $\pm$ 0.041 (0.630) & 0.512 $\pm$ 0.009 (0.497) & 0.354 $\pm$ 0.066 \textbf{(0.260)} \\
\midrule
\multirow{2}{*}{DESN S.} & IID & 0.643 $\pm$ 0.025 (0.618) & 0.734 $\pm$ 0.050 (0.690) & 0.631 $\pm$ 0.021 (0.608) & 0.684 $\pm$ 0.049 (0.600) \\
 & ON & 0.635 $\pm$ 0.022 (0.612) & 0.708 $\pm$ 0.012 (0.690) & 0.562$^{*}$ $\pm$ 0.022 (0.545) & 0.458$^{*}$ $\pm$ 0.056 (0.360) \\
\midrule
\multirow{2}{*}{DESN S.F.} & IID & 0.666 $\pm$ 0.031 (0.604) & 0.752 $\pm$ 0.026 (0.710) & 0.620 $\pm$ 0.015 (0.602) & 0.618 $\pm$ 0.053 (0.540) \\
 & ON & 0.639 $\pm$ 0.027 (0.608) & 0.716 $\pm$ 0.063 (0.620) & 0.558$^{*}$ $\pm$ 0.008 (0.548) & 0.460$^{*}$ $\pm$ 0.078 (0.310) \\
\midrule
\multirow{2}{*}{DESN C.} & IID & 0.644 $\pm$ 0.011 (0.628) & 0.736 $\pm$ 0.085 (0.620) & 0.633 $\pm$ 0.021 (0.614) & 0.666 $\pm$ 0.048 (0.600) \\
 & ON & 0.663 $\pm$ 0.014 (0.636) & 0.724 $\pm$ 0.030 (0.680) & 0.561$^{*}$ $\pm$ 0.011 (0.550) & 0.482$^{*}$ $\pm$ 0.057 (0.420) \\
\bottomrule
\end{tabular}%
}
\end{table*}
\vspace{-1 cm}
\subsubsection{ESNs excel at memorization.} Regarding ESN and LESN, our evaluations show that the standard ESN combined with an ON initialization is vital for successfully resolving the Copy task. Conversely, the memory capacity of LESN is neutral to the choice of input initialization, aligning with similar theoretical results \cite{ballarin2024memory}. 

We provide details on memorization of ESNs, let's take the state tracking tasks, where the target follows the form $y_t = u_{t-\Delta(t)}$. Consider that the input values during the delay ($\Delta$) and following the trigger step are uniformly zero. Consequently, the LESN enters a purely autonomous dynamical regime governed by $h_t = h_{t-1} W_{res}$, and also the ESN, as the non-linear $\tanh$ activation tends to operate in this regime because it has a stable fixed point at zero.

This quasi-linear regime is strongly encouraged by ON initialization rather than IID initialization, as the orthonormal transformation preserves state magnitudes and prevents early neuron saturation ($ \tanh(x) \approx x $ for $x \approx 0$). Then, to understand why these tasks become solvable under autonomous reservoir states, we analyze it through the lens of linear memory capacity \cite{jaeger2001short}. This corresponds to the rank of the accumulated Krylov matrix: $\text{MC} = \text{rank}(\mathcal{K}_t) \leq d_r$ \cite{ballarin2024memory}. This matrix naturally appears by unrolling the LESN recurrence: $h_t = \sum_{k=0}^{t-1} u_{t-k} W_{in} W_{res}^{k+\Delta(t)}.$

The Krylov matrix is defined as follows: $\mathcal{K}_t = \left[ W_{in}, W_{in} W_{res}, \dots , W_{in} W_{res}^{t-1}  \right] $. The complete reservoir history of these tasks is contained within the Krylov subspace spanned by the columns of $\mathcal{K}_t^{\Delta} = \mathcal{K}_t W_{res}^{\Delta}$. Then, we require the readout to satisfy: $U \mathcal{K}_t^{\Delta} W_{out}= y_t$, where $U=[u_0,u_1,...]$. Therefore, the readout must invert the reservoir dynamics to isolate the specific delay. A solution $W_{out}$ with zero error exists only if the Krylov matrix $\mathcal{K}_t^{\Delta}$ preserves the information, that is, $\text{rank}(\mathcal{K}_t^{\Delta})= min(d_r, t \cdot d_i)= t\cdot d_i$, or $d_r \ge t \cdot d_i$, ensuring that distinct input histories map to distinct reservoir states. Note that the rank condition is invariant to the delay, since $W_{res}$ is non-singular $\text{rank}(\mathcal{K}_t^{\Delta})= \text{rank}(\mathcal{K}_t)$. Nevertheless, in practice, the system suffers from a memory gap or Krylov subspace squeezing. This phenomenon manifests as the ill-conditioning of the Gram matrix required for the pseudoinverse in \eqref{eq:pseudoinverse}, caused by the rotations $W_{in} W_{res}^{j}$ becoming collinear rapidly. This effect is significantly less pronounced when $W_{res}$ is orthogonal \cite{ballarin2024memory,white2004short}.

Regarding LeakyESN, on one hand, hyperparameter validation showed that the optimal leak rate across most state tracking tasks is $\alpha = 0.9$, which closely approximates a standard ESN. In this manner, LeakyESN has also the capacity of solving the Copy task with ON initialization. On the other hand, LeakyESN IID found lower leak rates ($\alpha \in [0.1, 0.3]$) and achieved the best performance among reservoir models on the Sorting task. A task where LESN and ESN fail. This indicates that while the contribution of the nonlinear activation may be constrained by low leak rates, its presence remains structurally necessary for some complex tasks.

\subsubsection{How G-ESN solves tasks unattainable by other ESNs.} 
\begin{figure}[h]
\centering
  \includegraphics[scale=0.3]{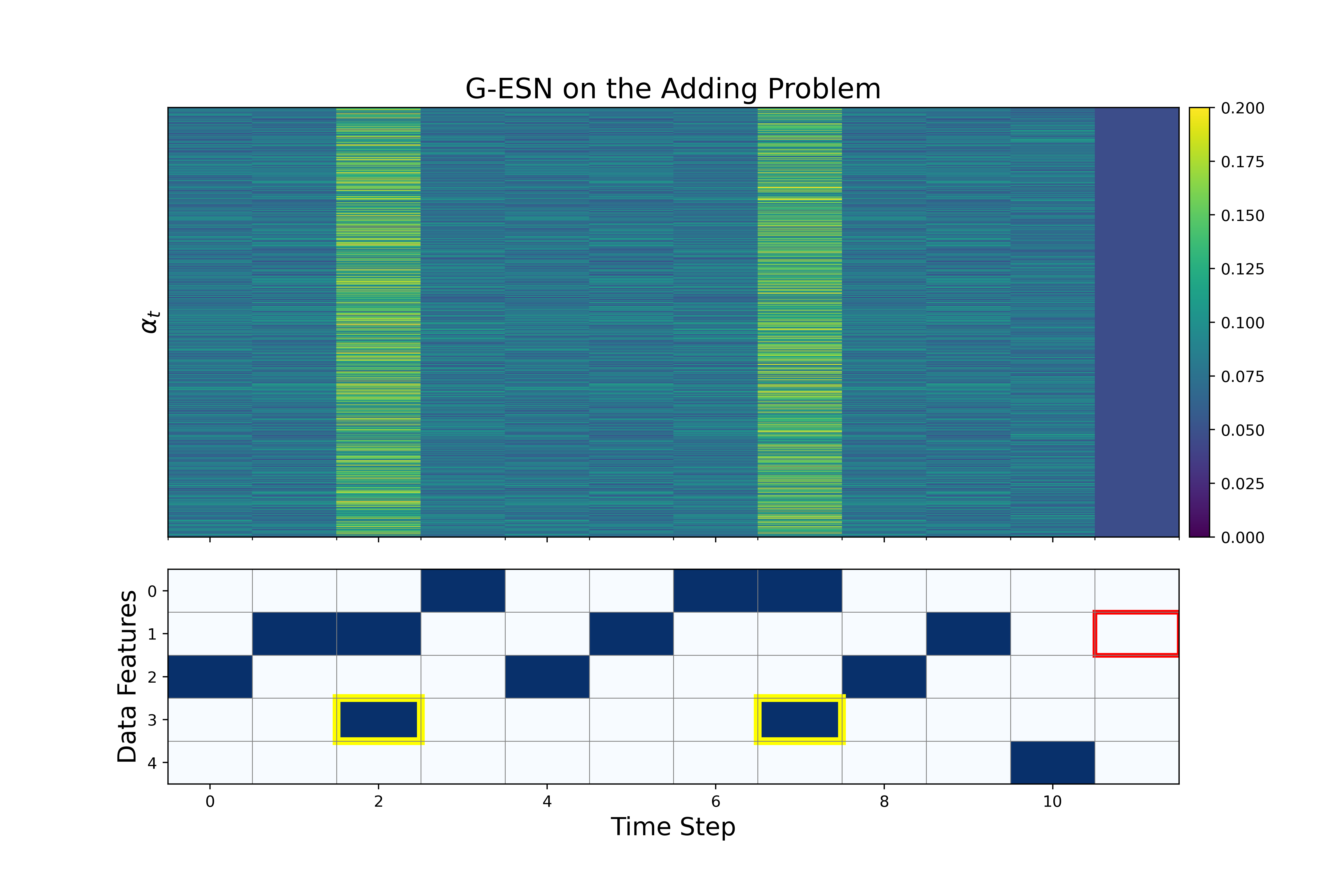} 
  \vspace{-0.5 cm}
  \caption{We represent a sample of the adding problem, which is a sequence of $12$ elements: a sequence of $10$ time steps with three symbols (first three rows of Data Features), two markers (fourth row of Data Features) in yellow (at time steps $2$ and $7$), a trigger (last row of Data Features), and a last state expecting the result (marked by the red box). We must learn that the sum of $1$ (at $t=2$) plus $0$ (at $t=7$) equals $1$ (at $t=11$). Note that $\alpha_t$ has $d_r=2000$ as $d_o=5$ possible results.}
  \label{fig:adding_gatedESN}
\end{figure}

The G-ESN ON achieved the top performance among reservoir models on SMNIST and S-copy (this with IID as well). G-ESN IID successfully solved Adding, where all other reservoir models failed, showing near-perfect accuracy with minimal variance. The data-dependent leaky integrator functions as an adaptive statistical filter, effectively emulating an attention mechanism. During hyperparameter validation, the optimal gating bias was found to be $b = -3$, resembling the analysis presented, under this configuration, standard symbols (lacking active markers) fail to open the gate, whereas marked tokens overcome the negative bias to update the reservoir state. A visualization of this sequential filtering mechanism is illustrated in Fig. \ref{fig:adding_gatedESN}.

ON initialization provided no statistically significant improvement over IID initialization for the bracket task across any of the ESN variants. Notably, G-ESN is also the unique reservoir model able to solve this task. From the mathematical description of the Bracket task, we deduce that the reservoir primarily needs to accumulate information over time to track nesting depth ($d_t$ in Appendix Sec. 3.12). The validation results showed that the optimal gating bias for this task was $b_{gate} = 0$, resembling the analysis presented that with our contrastive initialization for paired symbols (e.g., $($ and $)$). Mapping complementary symbols to opposite signed domains is crucial, removing this contrastive initialization degraded performance \footnote{G-ESN results without the restriction: $0.324 \pm 0.090 (0.160)$  for IID and $0.378 \pm 0.098 (0.250)$ for ON. We did not observe that this benefit was able to be transferred from $W_{gate}$ to $W_{in}$ for the other methods, possibly destroy by $W_{res}$.}. In conclusion, even when initialized with random weights, a properly tuned leaky integrator allows a sub-population of reservoir neurons to act as an statistical filter.
\vspace{-0.9 cm}
\vspace{-1.2 cm}
\subsubsection{Effect of structural layer consensus in Deep ESNs.} Similarly, ON initialization significantly improves the performance of Deep ESNs. For example, using the DESN S. method, switching from IID to ON initialization reduced errors on SMNIST (from $0.684$ to $0.458$), Disc. patt. (from $0.145$ to $0.060$), Copy (from $0.479$ to $0.000$), and Sorting (from $0.631$ to $0.562$).

Hyperparameter validation confirmed that the shared readout architecture requires higher regularization ($\lambda \in [1, 10]$) to enforce feature invariance and consensus across layers. In contrast, the conventional concatenated readout optimizes under low regularization ($\lambda = 0.0001$), encouraging independent feature selection from isolated layers within the expanded state space. This distinction is particularly evident in the bracket task under ON initialization. DESN S.F. yields marginal performance improvements over both DESN C. DESN S. variants on SMNIST, Adding, S-copy, and Bracket tasks, while remaining competitive across all other tasks.

\begin{table*}[t!]
    \centering
    \resizebox{\textwidth}{!}{%
    \begin{threeparttable}
        \caption{Performance comparison on the Stream benchmark. Rows are color-coded to group tasks where the gap between SOTA and ESN is less than $0.05$. }
        \label{tab:best_performance}
        \definecolor{grupo1}{RGB}{255, 236, 237}
        \definecolor{grupo2}{RGB}{229, 242, 255}  
        \definecolor{grupo3}{RGB}{229, 255, 204} 
        \begin{tabular}{l c cc c cc}
            \toprule
            & & \multicolumn{2}{c}{\textbf{SOTA}} & & \multicolumn{2}{c}{\textbf{ESN}} \\
            \cmidrule{2-4} \cmidrule{6-7}
            \textbf{Task} & \textbf{S6} & \textbf{Best \cite{bendi2025echo}} & \textbf{Model \cite{bendi2025echo}} & & \textbf{Best} & \textbf{Model}\\
            \midrule
            \rowcolor{grupo1} S-Copy            & \textbf{0.348} & 0.426          & EST / 10k          & & 0.429          & G-ESN / IID \\
             \rowcolor{grupo1} Sorting           & \textbf{0.358} & 0.473          & EST / 10k          & & 0.477          & LeakyESN / IID \\
             \midrule
             \rowcolor{grupo2} Sinus    & 0.009 & 0.007          & Transformer / 100k & & \textbf{0.000} & ESN / IID \\    
             \rowcolor{grupo2}  Chaotic       & 0.028 & 0.011          & Transformer / 1M   & & 0.011          & ESN / IID \\        
             \rowcolor{grupo2} Disc. post.       & 0.432 & \textbf{0.000} & EST \& GRU / 10k   & & \textbf{0.000} & \tnote{1} \\
             \rowcolor{grupo2} Cont. post.    & 0.167 & 0.005          & EST / 1k           & & \textbf{0.000} & LESN / IID \& ON \\
             \rowcolor{grupo2} Cont. patt.  & 0.057 & 0.074          & EST / 1k           & & \textbf{0.023}          & ESN / IID \\
             \rowcolor{grupo2} Adding     & \textbf{0.000} & \textbf{0.000} & Transformer / 10k & & 0.006          & G-ESN / IID \\

             \rowcolor{grupo2} SMNIST          & 0.430 & \textbf{0.328} & Transformer / 1M   & & 0.348          & G-ESN / ON \\
             \midrule
             \rowcolor{grupo3} Disc. patt.   & 0.328 & 0.166          & EST / 10k          & & \textbf{0.056} & LESN / IID \\
                        
             \rowcolor{grupo3} Bracket   & 0.260 & 0.232 & EST / 1k           & & \textbf{0.040} & G-ESN / IID \& ON \\
             \rowcolor{grupo3} Copy               & 0.515 & 0.541          & EST / 1k           & & \textbf{0.000} & \tnote{2}\\
            \bottomrule
        \end{tabular}
        
        \begin{tablenotes}
            \scriptsize 
            \item[1] DESN C. / IID \& ON - DESN S. / ON - DESN S.F. / IID \& ON - ESN / ON - LeakyESN / ON - LESN / ON.
            \item[2] DESN S.F. / ON - ESN / ON - LESN / IID \& ON. 
        \end{tablenotes}
    \end{threeparttable}
    }
\end{table*}

\subsubsection{Comparison against SOTA models.}  Tab. \ref{tab:best_performance} compares SOTA and our ESNs models. ESN variants prove competitive across most of these few-shot seq2seq pattern tasks. The primary exceptions are S-copy, Sorting, and SMNIST, where S6 excels on the first two tasks and Transformers dominate the latter. Indeed, S6 was proposed to solve the former, as a content-aware task \cite{gu2024mamba}. 

These ESN results are noteworthy considering that they employ an efficient closed-form, mean-squared-error–based training procedure even for classification, whereas the SOTA models rely on extensive optimization using backpropagation guided by cross-entropy loss. Furthermore, ESNs demonstrate superiority over SOTA models on the Disc. patt., Bracket and Copy tasks. 

\begin{figure}[h!]
\centering
  \includegraphics[scale=0.3]{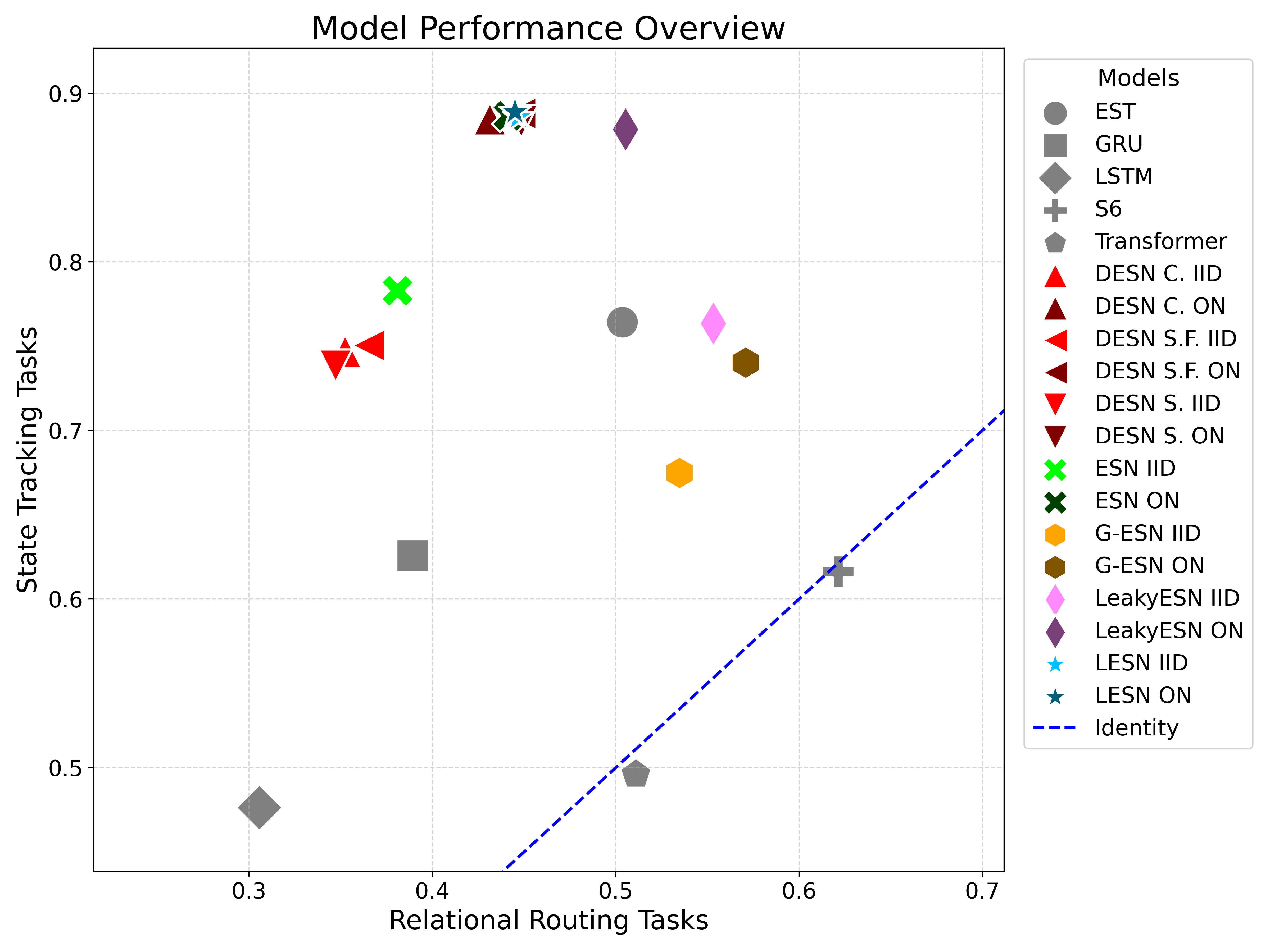} 
  \caption{Accuracy comparison between two groups of task (high is better). Colors differentiate IID and ON initialization methods. Identity denotes the function $y=x$.}
  \label{fig:comparison}
  \vspace{-0.55 cm}
\end{figure}

Fig. \ref{fig:comparison} summarizes these architectural trade-offs. The ESN variants cluster primarily above the identity line, showing a strong inductive bias toward state tracking tasks over relational routing tasks. Specifically, the ON initialization for ESN, LeakyESN, Deep ESNs, and LESN (IID and ON) achieve the highest overall accuracy on state tracking tasks. Conversely among reservoir models, the G-ESN ON and LeakyESN IID architectures perform better on relational routing tasks. Finally, gradient-based methods like S6 and Transformers cluster below the identity line, 
confirming their strength in relational routing tasks, though often accompanied by a performance trade-off on pure state-tracking tasks.

\section{Conclusion}
\vspace{-0.5 cm}
This paper presented an empirical evaluation of ESNs evaluated on few-shot seq2seq pattern tasks. As summarized in the introduction, our results reveal relationships and takeaways between specific reservoir properties and task-level demands. A limitation of this study stems from the fixed evaluation protocol of the STREAM benchmark. Future work could extend these assessments by varying sequence lengths, number of symbols, data partitions and seeds. Although the G-ESN model with properly tuned distributional parameters demonstrates the advantages of a proper initialization of the leak rate, its generalization to other tasks is not straightforward since these parameters are not automatically learned.

Next steps will focus on exploring how our insights apply to content-aware tasks in natural language processing, investigate potential synergies with modern linear-attention frameworks, and analyze how distributional, architectural, and optimization ESN variants behave when processing continuous positional encodings rather than discrete one-hot vectors.
\vspace{-0.1 cm}
\begin{credits} 
\subsubsection{\ackname} Supported by the 'Siemens Energy AI Chair: Energy Sustainability for a Decarbonized Society 5.0' (TSI-100930-2023-5), funded by the Secretary of State for Digitalization and Artificial Intelligence through the ENIA 2022 Chairs call, and co-funded by the European Union-Next Generation EU. And supported by the Spanish research grant (PID2022-143299OB-I00/AEI/
10.13039/501100011033/FEDER, UE [A.T.-D., E.R., and A.V.]).
\end{credits}
\vspace{-0.1 cm}

\bibliographystyle{splncs04}
\bibliography{mybibliography}

\end{document}